\documentclass[sigconf]{acmart}
\AtBeginDocument{%
  }

\copyrightyear{2025}
\acmYear{2025}
\setcopyright{acmlicensed}
\acmConference[MM '25]{Proceedings of the 33rd ACM International Conference on Multimedia}{October 27--31, 2025}{Dublin, Ireland}
\acmDOI{10.1145/3746027.3754993}
\acmISBN{979-8-4007-2035-2/2025/10}

\usepackage{multirow}
\usepackage{booktabs}
\usepackage{array} 
\usepackage{arydshln}
\usepackage{makecell}

\begin{document}

\title[MIHBench \& DAB]
      {MIHBench: Benchmarking and Mitigating Multi-Image Hallucinations in Multimodal Large Language Models}


\author{Jiale Li}
\email{jialeli@stu.xmu.edu.cn}
\affiliation{
  \institution{Xiamen University}
  \department{Key Laboratory of Multimedia Trusted Perception and Efficient Computing, Ministry of Education of China}
  \city{Xiamen}
  \state{Fujian}
  \postcode{361005}
  \country{China}
}

\author{Mingrui Wu}
\email{mingrui0001@gmail.com}
\authornote{Equal contribution to Jiale Li.}
\affiliation{
  \institution{Xiamen University}
  \department{Key Laboratory of Multimedia Trusted Perception and Efficient Computing, Ministry of Education of China}
  \city{Xiamen}
  \state{Fujian}
  \postcode{361005}
  \country{China}
}
\affiliation{
  \institution{Zhongguancun Academy}
  \city{Beijing}
  \country{China}
}

\author{Zixiang Jin}
\email{37220222203643@stu.xmu.edu.cn}
\affiliation{
  \institution{Xiamen University}
  \department{Key Laboratory of Multimedia Trusted Perception and Efficient Computing, Ministry of Education of China}
  \city{Xiamen}
  \state{Fujian}
  \postcode{361005}
  \country{China}
}

\author{Hao Chen}
\email{37120222203278@stu.xmu.edu.cn}
\affiliation{
  \institution{Xiamen University}
  \department{Key Laboratory of Multimedia Trusted Perception and Efficient Computing, Ministry of Education of China}
  \city{Xiamen}
  \state{Fujian}
  \postcode{361005}
  \country{China}
}

\author{Jiayi Ji}
\email{jjyxmu@gmail.com}
\authornote{Corresponding author.}
\affiliation{
  \institution{Xiamen University}
  \department{Key Laboratory of Multimedia Trusted Perception and Efficient Computing, Ministry of Education of China}
  \city{Xiamen}
  \state{Fujian}
  \postcode{361005}
  \country{China}
}

\author{Xiaoshuai Sun}
\email{xssun@xmu.edu.cn}
\affiliation{
  \institution{Xiamen University}
  \department{Key Laboratory of Multimedia Trusted Perception and Efficient Computing, Ministry of Education of China}
  \city{Xiamen}
  \state{Fujian}
  \postcode{361005}
  \country{China}
}

\author{Liujuan Cao}
\email{caoliujuan@xmu.edu.cn}
\affiliation{
  \institution{Xiamen University}
  \department{Key Laboratory of Multimedia Trusted Perception and Efficient Computing, Ministry of Education of China}
  \city{Xiamen}
  \state{Fujian}
  \postcode{361005}
  \country{China}
}

\author{Rongrong Ji}
\email{rrji@xmu.edu.cn}
\affiliation{
  \institution{Xiamen University}
  \department{Key Laboratory of Multimedia Trusted Perception and Efficient Computing, Ministry of Education of China}
  \city{Xiamen}
  \state{Fujian}
  \postcode{361005}
  \country{China}
}

\renewcommand{\shortauthors}{Jiale Li and Mingrui Wu et al.}

\begin{abstract}
  Despite growing interest in hallucination in Multimodal Large Language Models (MLLMs), existing studies primarily focus on single-image settings, leaving hallucination in multi-image scenarios largely unexplored. To address this gap, we conduct the first systematic study of hallucinations in multi-image MLLMs and propose \textbf{MIHBench}, a benchmark specifically tailored for evaluating object-related hallucinations across multiple images. MIHBench comprises three core tasks—Multi-Image Object Existence Hallucination, Multi-Image Object Count Hallucination, and Object Identity Consistency Hallucination—targeting semantic understanding across object existence, quantity reasoning, and cross-view identity consistency. Through extensive evaluation, we identify key factors associated with the occurrence of multi-image hallucinations, including: (1) a progressive relationship between the number of image inputs and the likelihood of hallucination occurrences; (2) a strong correlation between single-image hallucination tendencies and those observed in multi-image contexts; and (3) the influence of same object image ratios and the positional placement of negative samples within image sequences on the occurrence of object identity consistency hallucination. To address these challenges, we propose a \textbf{Dynamic Attention Balancing~(DAB)} mechanism that adjusts inter-image attention distributions while preserving the overall visual attention proportion. Experiments across multiple state-of-the-art MLLMs demonstrate that our method effectively reduces hallucination occurrences and enhances semantic integration and reasoning stability in multi-image scenarios. The project page is available at~\url{https://github.com/pgtrece/DAB/}.
\end{abstract}



\maketitle

\section{Introduction}

In recent years, integrating vision encoders~\cite{clip} with Large Language Models (LLMs)~\cite{qwenllm} has driven significant progress in multimodal large language models (MLLMs) like LLaVA-1.5~\cite{llava1.5} and others~\cite{minigpt4,shikra,owl2,blip2,chatglm,cmp,rstnet}, achieving strong results in tasks such as visual question answering and vision-language reasoning~\cite{kosmos,few_shot,llavamed,RCOD,TAR3D,AR,gu2023orsi,gu2025acl,gu2025optical}. However, most focus on single-image inputs. To meet growing demands for richer semantic understanding, multi-image processing has emerged, enabling extraction of more diverse visual information. For example, Qwen-VL 2.5~\cite{qwen2.5vl} supports multi-image inputs to improve context comprehension. Correspondingly, new benchmarks~\cite{milebench,micbench,mmdu,nlvr}, including MMIU~\cite{mmiu} and MuirBench~\cite{muirbench}, have been developed to evaluate multi-image reasoning across various tasks.

\begin{figure*}[t]
  \centering
  \includegraphics[width=\textwidth]{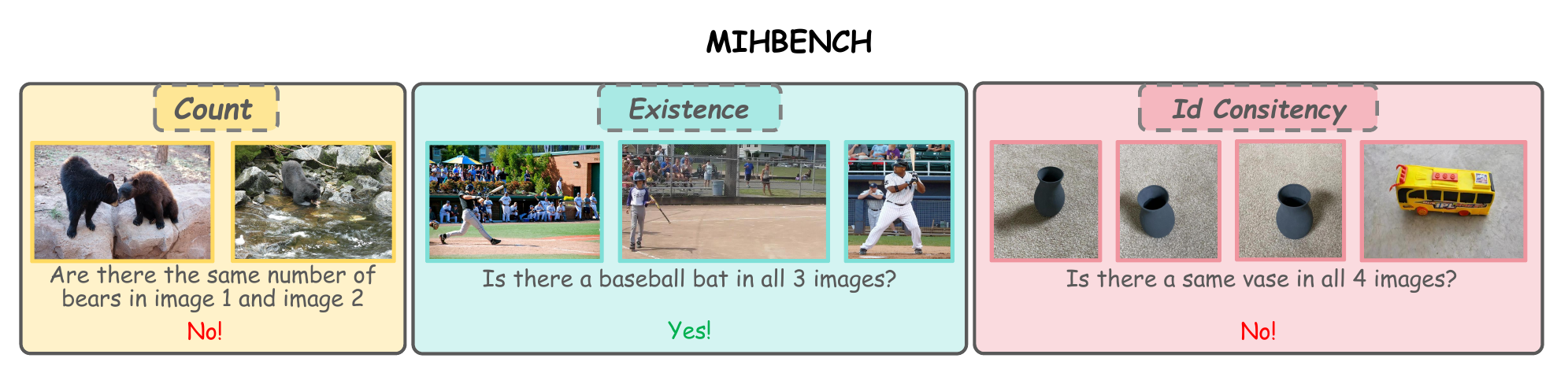}
  \caption{The overview of our proposed MIHBench. MIHBench consists of three distinct categories: multi-image object count hallucination, multi-image object existence hallucination, and object identity consistency hallucination. }
  \label{ourbench}
\end{figure*}

Despite strong performance on general benchmarks, growing evidence reveals that MLLMs often produce outputs misaligned with the visual input—a phenomenon known as multimodal hallucination~\cite{survey1,survey2,rbench,difnet,lure,halucidoctor}. Due to the inherent challenges in diagnosing and mitigating this issue, many studies~\cite{pai,vcd,opera,cca,see} have focused on understanding and reducing hallucinations in MLLMs. Nevertheless, these efforts have largely concentrated on models with single-image input capabilities, leaving hallucination in the context of multi-image MLLMs underexplored.

To address this gap, we present \textbf{MIHBench}, the first dedicated benchmark designed to evaluate hallucination phenomena in multi-image MLLMs. MIHBench comprises three representative tasks: Multi-Image Object Existence Hallucination, Multi-Image Object Count Hallucination, and Object Identity Consistency Hallucination. Through comprehensive evaluation on several state-of-the-art multi-image MLLMs, we observe three major factors that contribute to hallucination in this context: (1) hallucination frequency increases with the number of input images, indicating a deficiency in semantic integration across images; (2) hallucination tends to propagate from one image to others, highlighting a contagion effect from single-image misperceptions; and (3) the position of distractor images within the sequence significantly impacts model performance, with later-positioned distractors more likely to be overlooked. 

To mitigate these issues, we propose a lightweight \textbf{Dynamic Attention Balancing (DAB)} mechanism that regulates the distribution of attention across images during decoding.
Our method adaptively balances the attention weights of image tokens without introducing extra inference overhead. 
By applying such soft constraints that ensure each image receives an approximately equal share of attention, our method significantly reduces hallucination in MLLMs. Experiments across representative MLLMs demonstrate consistent improvements on all three MIHBench tasks, confirming the generalizability and effectiveness of our approach for alleviating hallucinations in multi-image reasoning scenarios.

In summary, our contributions are as follows:

\begin{itemize}
    \item We introduce \textbf{MIHBench}, the first benchmark explicitly designed to evaluate hallucination in multi-image MLLMs. It includes three complementary tasks—multi-image object existence hallucination, multi-image object count hallucination, and object identity consistency hallucination—capturing different facets of multi-image hallucination behaviors.
    
    \item We conduct the first systematic study of hallucination in multi-image settings. Our experiments reveal that hallucination severity increases with more input images, can propagate from one image to others, and is highly sensitive to the position of distractor images. These insights shed light on the unique challenges posed by multi-image inputs.

    \item We propose a \textbf{Dynamic Attention Balancing (DAB)} mechanism to mitigate multi-image hallucinations. DAB adaptively rebalances attention weights across image tokens in a lightweight and training-free manner, leading to consistent improvements across various models and tasks.
\end{itemize}

\section{Related Work}

\subsection{Multimodal Large Language Models}

MLLMs have evolved from single-image tasks like captioning and VQA~\cite{llava, iblip} to complex multi-image and temporal scenarios. Early models faced challenges in cross-image reasoning due to limited visual token processing and inter-image semantic modeling. Recent works~\cite{idefics,idefics2,xcomposer2,flamingo,deepseekvl,longllava,deepseekvl2,emu} tackle these via architectural and data innovations. Notably, Qwen-VL 2.5 enhances sequential understanding with dynamic token modulation~\cite{qwen2.5vl}; LLaVA-NeXT-Interleave unifies diverse inputs~\cite{llava_next_internleave}; Mantis leverages large interleaved instruction tuning~\cite{manits}; InternVL 2.5 handles high-res multi-image/video inputs, setting new MMMU benchmarks~\cite{mmmu,internvl2.5}. These advances improve MIQA, storytelling, and multi-view inference, advancing MLLM cognition.

\subsection{Multimodal Hallucination Benchmarks}

Multimodal hallucination—textual outputs inconsistent with visual inputs—is categorized into object, relational, and attribute hallucinations. Existing benchmarks~\cite{hallusionbench,rbench,eventhal,reefknot,longhal,mhalubench,rope} mainly focus on single-image settings. For instance, CHAIR~\cite{chair} quantifies object hallucinations via hallucinated entities in captions; POPE~\cite{pope} uses voting-based object presence evaluation; R-Bench targets relational hallucinations at image and object levels. However, these do not cover multi-image hallucination. We propose MIHBench, the first benchmark for evaluating hallucinations in multi-image contexts, advancing multi-image reasoning evaluation.

\subsection{Hallucination Mitigation in MLLMs}

Various hallucination mitigation methods~\cite{cca,ibd,agla,pai,controlmllm,icd} mostly avoid additional training. OPERA~\cite{opera} penalizes over-reliance on specific tokens during decoding with rollback strategies; VCD~\cite{vcd} contrasts logits between original and distorted images to reduce bias; Woodpecker~\cite{woodpecker} applies a five-stage post-hoc correction pipeline including concept extraction and visual verification. These methods focus on single-image inputs and incur high costs when extended to multi-image scenarios. We propose a novel approach that dynamically balances attention across multiple images with minimal inference overhead, effectively reducing hallucinations in multi-image inputs.

\section{MIHBench}

This section introduces MIHBench, the first benchmark specifically designed to evaluate object-level visual hallucination in multi-image MLLMs. As illustrated in Figure~\ref{ourbench}, the benchmark encompasses three evaluation tasks: \textit{multi-image object existence hallucination}, \textit{multi-image object count hallucination}, and \textit{object identity consistency hallucination}. The MIHBench dataset comprises a total of 3,527 images and 4,000 questions, as shown in Table~\ref{mihbench_num}.  Detailed construction workflows for each task are provided in the the supplementary materials.

\subsection{Multi-Image Object Existence Hallucination}

The \textit{multi-image object existence hallucination} task aims to assess whether a model can accurately determine the presence of a specific object across multiple images. We adopt a POPE-liked~\cite{pope} voting mechanism for evaluation. The question template is: 

\textbf{``Is there a/an <object> in all 3 images?''} 

A ``Yes'' response indicates that the model believes the object appears in all input images; a ``No'' response suggests that the object is absent from at least one of the images. This task necessitates comprehensive understanding of each image and the ability to aggregate visual cues across all three.

\noindent
\textbf{Data Construction:} We construct a rigorous evaluation framework for object existence hallucination across multiple images. We first annotate the MSCOCO2014~\cite{mscoco2014} validation set utilizing the Grounding SAM model~\cite{gdsam} to establish reliable ground truth annotations. For each image, we extract all object categories defined in the COCO dataset along with their corresponding confidence scores. Objects with confidence scores below 0.5 are considered absent from the image, thereby ensuring a conservative approach to annotation.

Following annotation, we adopt the taxonomic structure established in POPE to construct three distinct question-answer (QA) categories: \textit{random}, \textit{popular}, and \textit{adversarial}. Based on these QA pairs, we identify images containing the same queried object and randomly sample three images per object group to form multi-image evaluation tuples.

Each question sample is formally represented as: 
\[
\langle [x_1, x_2, x_3], q(o), a = l_1 \land l_2 \land l_3 \rangle,
\]
where $x_i$ denotes the $i$-th image, $q(o)$ represents the question regarding object $o$, and $l_i \in \{0,1\}$ indicates the ground truth label for image $x_i$. The final ground truth answer $a$ is computed as the logical conjunction over all individual ground truths, ensuring the task accurately evaluates multi-image object existence reasoning.

We construct 800 QA instances for each of the three subtypes, yielding a total of 2,400 questions with an equal distribution of positive and negative samples.

\begin{table}[t]
  \caption{Statistics of images and questions in the MIHBench. The dataset comprises a total of 3,527 images and 4,000 voting-based binary (yes/no) questions across three task types: \textit{Existence}, \textit{Count}, and \textit{Identity Consistency}. }
  \label{mihbench_num}
  \begin{tabular}{lcc}
    \toprule
    Type & Image Count & Question Count \\
    \midrule
    Existence & 500  & 2400 \\
    Count & 1440 & 800 \\
    Identity Consistency & 1619 & 800 \\
    \bottomrule
  \end{tabular}
\end{table}

\subsection{Multi-Image Object Count Hallucination}

The \textit{multi-image object count hallucination} task is designed to evaluate a model's ability to accurately compare the counts of a specific object category across two images. Similar to the previous task, we adopt a voting-based evaluation approach. The question template used is as follows:

\textbf{``Are there the same number of <object> in all 2 images?''}

A ``Yes'' response indicates that the model believes the target object appears in equal quantity across all input images, whereas a ``No'' response indicates a perceived discrepancy in object counts. This task demands fine-grained object-level understanding and the ability to reason about inter-image attribute consistency, particularly with respect to object quantities.

\begin{figure}[t]
  \centering
  \includegraphics[width=0.9\linewidth]{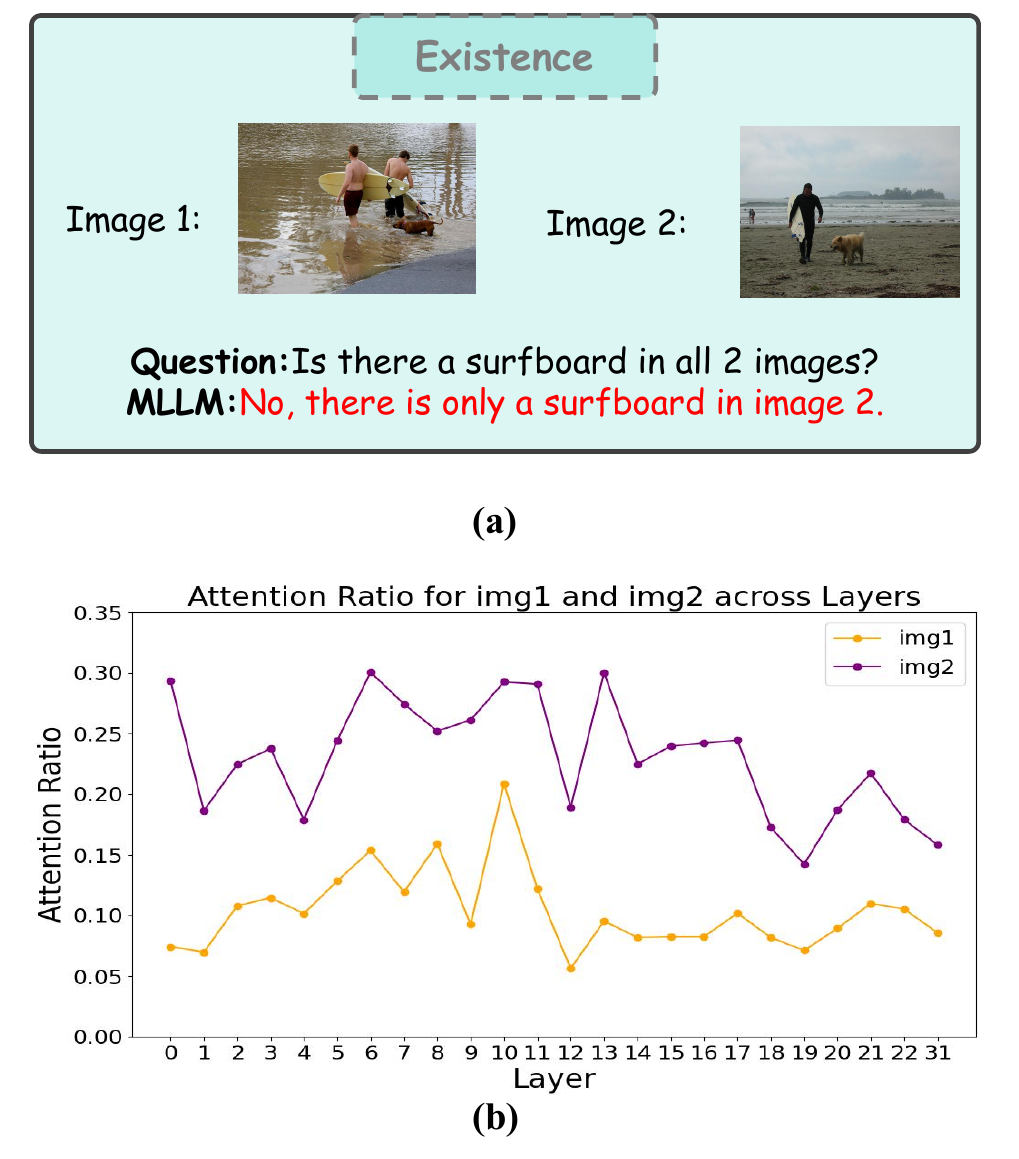}
  \caption{(a) The MLLM incorrectly concludes that only the Image2 contains a surfboard, fails to recognize an actually present entity in Image1.
(b) The average attention ratio across layers shows that Image 1 receives consistently less attention than Image 2. 
This shows that the imbalanced attention distribution induces multi-image hallucination.}
  \label{method_start}
\end{figure}

\noindent
\textbf{Data Construction:} We employ a similar annotation methodology using the Grounding SAM model on the MSCOCO 2014~\cite{mscoco2014} validation set, but with a focus on extracting accurate object counts per category per image. To prevent object overcrowding within a single image from impairing the model’s ability to accurately count and compare object quantities, we constrain the candidate object categories to those with no more than three instances per image. We count only instances with confidence scores of 0.5 or higher, ensuring that enumerated objects possess sufficient visual saliency to be reliably detected.

Based on these annotations, we formulate question samples as:
\[
\langle [x_1, x_2], q(o), a = \mathbb{I}[n_1 = n_2] \rangle,
\]
where $x_1$ and $x_2$ represent the image pair, $q(o)$ denotes the object-related question, $n_1$ and $n_2$ correspond to the object counts in each respective image, and $a$ represents the ground truth determined by count equality. Furthermore, to enhance evaluation robustness, we include image pairs lacking the queried object in one or both images during data construction.

Our constructed dataset comprises 800 question samples with balanced positive and negative instances. Notably, 200 positive examples feature image pairs where neither image contains the queried object, while 200 challenging negative examples include one image that lacks the queried object entirely.

\subsection{Object Identity Consistency Hallucination}

The \textit{object identity consistency hallucination} task is designed to evaluate a model’s ability to maintain object identity consistency across multiple images, particularly in the presence of distractor instances. Using a voting-based format, the question template is defined as:

\textbf{``Is there a same <object> in all 4 images?''}

A ``Yes'' response indicates that the model perceives the same object instance to be present in all input images, whereas a ``No'' response suggests that the model has identified an image that does not contain the same object, potentially a distractor. This task challenges the model’s robustness and consistency in recognizing object identity under multi-view conditions, even when unrelated objects are visually introduced.

\begin{figure}[t]
  \centering
  \includegraphics[width=\linewidth]{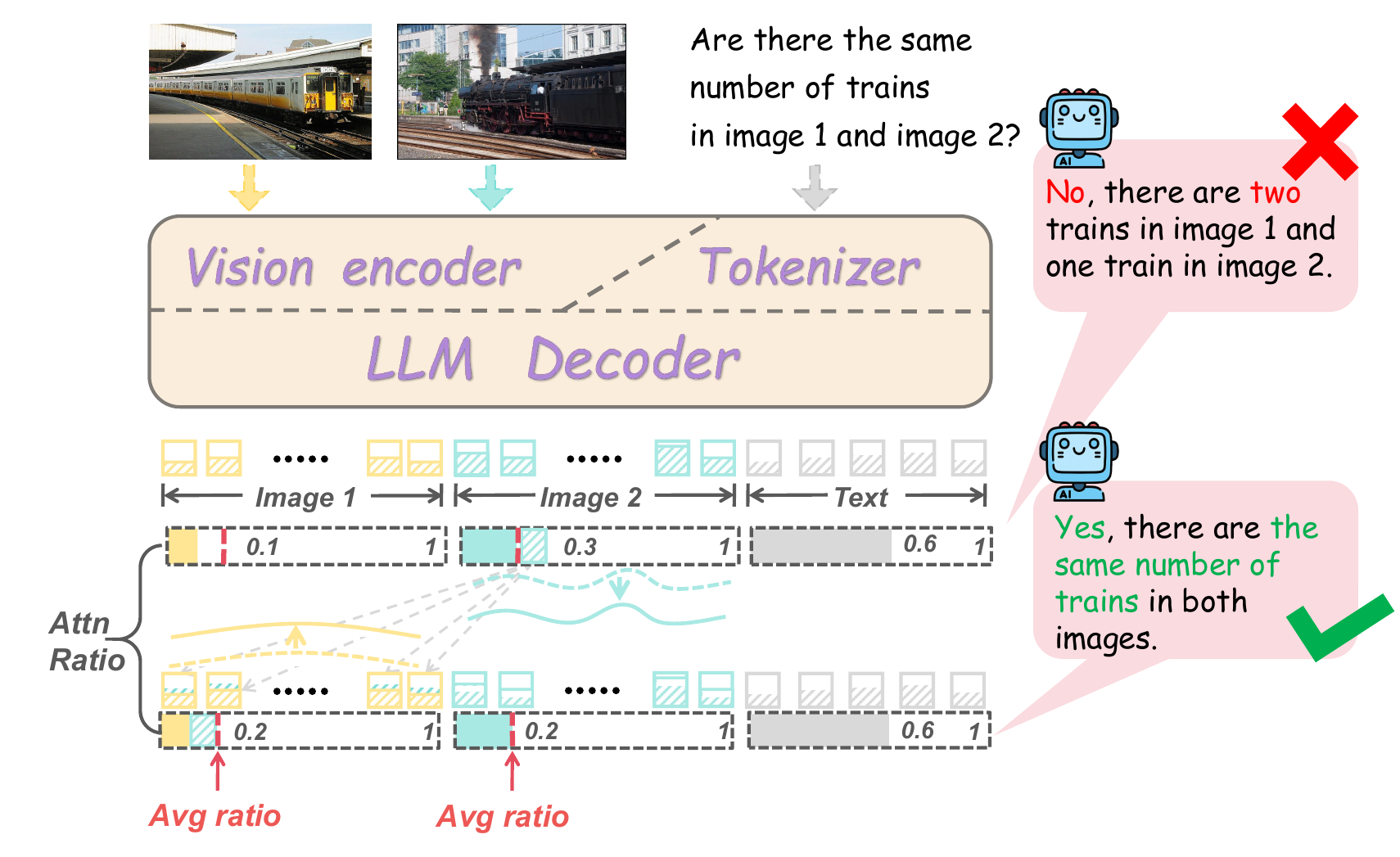}
  \caption{Dynamic Attention Balancing in Multi-Image VQA. We first compute the average attention ratio of image tokens across the input sequence. Then, for images whose attention exceeds the average, a portion of their attention is redistributed to under-attended images. This helps preserve the intra-image attention distribution structure while encouraging inter-image attention distributions to become more aligned. This dynamic balancing leads to more equitable attention allocation across images, mitigating hallucination.}
  \label{method}
\end{figure}

\noindent
\textbf{Data Construction:} We leverage the CO3D dataset~\cite{co3d} as the source of multi-view object images. CO3D consists of video sequences capturing 50 common COCO object categories, with each sequence documenting real-world objects from multiple viewpoints during complete 360-degree rotations.

For positive examples, we uniformly sample four frames from a single object's video sequence, representing distinct viewpoints of the same object instance. For negative examples, we employ a controlled contrastive approach: we randomly select three viewpoints of a target object and then, using CLIP~\cite{clip} similarity scores, identify the most visually dissimilar image from a different object category to serve as a distractor. This distractor is randomly inserted among the three target views, rather than consistently positioned as the fourth image, to increase example diversity and evaluation difficulty.

Our final dataset consists of 800 question samples, comprising 400 positive and 400 negative examples, maintaining balanced class distribution for robust evaluation.

\section{Method}

\subsection{Preliminaries}

\noindent
\textbf{Input Composition.} The inputs to MLLMs can be broadly divided into two types: visual inputs and textual inputs.  We denote the input as $\boldsymbol{H}^0 = [\mathit{V}_1, \mathit{V}_2, \ldots, 
\mathit{V}_n,$ $\mathit{X}]$, where each $\mathit{V}$ represents the image tokens derived from an image, $\mathit{X}$ denotes the text tokens obtained after tokenization, and $\mathit{n}$ indicates the number of images.

\noindent
\textbf{Attention Mechanism} The text generation capabilities of MLLMs are primarily governed by the internal decoder architecture of the underlying Large Language Model (LLM). Given an initial input representation $\boldsymbol{H}^0$, the computation within each layer of the LLM decoder can be formalized as follows:
\begin{equation}
\boldsymbol{A}^l = \mathrm{softmax} \left( \frac{\mathbf{Q}^l(\boldsymbol{H}^{l-1}) \cdot \mathbf{K}^l(\boldsymbol{H}^{l-1})^T}{\sqrt{d_K}} \right),
\end{equation}
where $l \in \{1, \dots, L\}$ denotes the index of the current decoder layer (out of a total of $L$ layers). The matrices $\mathbf{Q}^l$, and $\mathbf{K}^l$ represent the query, key, and value projections at the $l$-th layer, respectively. The attention matrix $\boldsymbol{A}^l \in \mathbb{R}^{d_q \times d_k}$ determines the attention weights that are used to compute contextualized representations throughout the entire sequence.

In the context of multi-image MLLMs, the attention mechanism within the decoder plays a critical role in enabling effective multimodal fusion. Specifically, the attention computation can be decomposed into intra-modal (unimodal) attention and cross-modal interaction components. For a given decoder layer $l$, the attention between the $i$-th image and the textual modality is formalized as:
\begin{equation}
A^l_i=\mathrm{softmax} \left(\frac{\mathbf{Q}^l(V_i^{l-1})\mathbf{K}^l(X^{l-1})^T}{\sqrt{d_K}}
\right),
\end{equation}
\noindent where $V_i^{l-1}$ denotes the hidden representation of the $i$-th image at the $(l{-}1)$-th layer, and $X^{l-1}$ denotes the hidden representation in $\boldsymbol{H}^{l-1}$ corresponding to the textual input embedding. This formulation facilitates fine-grained, image-specific attention over the textual representation, allowing the decoder to perform cross-modal reasoning across multiple visual inputs.

After propagating through all $L$ decoder layers, the model yields the final hidden representation $\boldsymbol{H}^L$, which is subsequently used to autoregressively predict the next token in the output sequence.

\subsection{Dynamic Attention Balancing Mechanism}

To explore the multimodal hallucinations induced by multiple image inputs, we visualize the attention weights assigned by the model to each image. It is clearly observed that the attention distribution across layers is not uniform for different images. As shown in Figure \ref{method_start}, the attention for the first image is significantly lower than that for the second image across all layers. Consequently, the model fails to identify the surfboard in the first image. We hypothesize that the unbalance attention allocation across multiple images leads to the model's overemphasis on the information from one image, neglecting the information from others. Therefore, when the model attempts to gather global visual information, which requires attending to all input images jointly, the bias towards a single image causes hallucinations. More analysis of the method will be presented in the supplementary materials.

\noindent
 \textbf{Dynamic Attention Balancing.} Based on the observations above, we propose a hallucination mitigation method Dynamic Attention Balancing that ensures each image is allocated approximately equal attention, as shown in Figure~\ref{method}.

Given the attention weight $\boldsymbol{A}^l_k$ computed between text tokens and the tokens of the k-th image, we compute the attention ratio for the $\mathit{k}$-th image over the entire input sequence, defined as:

\begin{equation}
\text{ratio}_k = \sum_{i=1}^{N_{I_k}} a_{i,j}^{l,h}, \quad
\text{for } i = 1, \dots, N_{I_k};\quad j = 1, \dots, N_X,
\end{equation}
where $a_{i,j}^{l,h} \in \boldsymbol{A}^l_k$,  $\mathit{N}_{I_k}$ denotes the number of image tokens in the $\mathit{k}$-th image, $\mathit{N}_X$ represents the number of text tokens, and $\mathbf{a}_{i,j}^{l,h}$ indicates the attention weight from the $\mathit{i}$-th image token of the $\mathit{k}$-th image to the $\mathit{j}$-th text token in the $\mathit{l}$-th layer and $\mathit{h}$-th head.

Next, we compute an average attention ratio used for attention balance. Inspired by previous work~\cite{see}, we only consider those valid visual related attention heads with the $\sum_{k=1}^{\mathit{n}} \mathit{ratio}_k$ is larger than $0.2$. For these heads, we compute the normalized attention ratio as follows:

\begin{equation}
\mathit{avg\_ratio} = \frac{\sum_{k=1}^{\mathit{n}} \mathit{ratio}_k}{\mathit{n}}.
\end{equation}

For image tokens with attention weight above $\mathit{avg\_ratio}$, we reduce their attention weight; for those below $\mathit{avg\_ratio}$, we increase it accordingly. We introduce a balancing coefficient $\boldsymbol{\alpha}$ to control the intensity of this adjustment. The attention shift is defined as:

\begin{equation}
\boldsymbol{\Delta}_{k,j}^{l,h} = \mathit{avg\_ratio} - \mathit{ratio}_k,
\end{equation}

\begin{equation}
\tilde{\mathbf{a}}_{i,j}^{l,h} = \mathbf{a}_{i,j}^{l,h} + \boldsymbol{\alpha} \cdot \frac{\boldsymbol{\Delta}_{k,j}^{l,h}}{\mathit{N}_{I_k}}.
\end{equation}

We apply this adjustment to each eligible attention head. When $\boldsymbol{\Delta}_{k,j}^{l,h} > 0$, the corresponding image token's attention is increased; when $\boldsymbol{\Delta}_{k,j}^{l,h} < 0$, it is reduced accordingly. 

\textbf{The design goal of DAB is to suppress extreme imbalance in attention distribution at the ``macro'' level (across images), while fully preserving the self-attention’s capability to focus on key tokens at the ``micro'' level (within each image token).} This is achieved by uniformly adjusting the attention scores---either increasing or decreasing by the same amount---for all image tokens belonging to the same image, thereby maintaining the internal semantic structure of each image while promoting balanced attention allocation across multiple images. This dynamic attention balancing mechanism ensures a more proportionally fair allocation of visual attention across images, thereby alleviating hallucinations caused by overconfidence on any single image when performing multi-image reasoning with MLLMs.

\begin{table*}[t]
\caption{
  Performance on the MIHBench benchmark. We compare baseline models and their DAB-enhanced variants across the three proposed tasks: multi-image object ex-
istence hallucination~(Existence), multi-image object count hallucina-
tion~(Count), and object identity consistency hallucination~(Identity Consistency). For the Existence task, results are averaged over the random, popular, and adversarial subsets, the specific results for each subtask will be presented in the supplementary materials. Our DAB method consistently improves performance across all tasks and models, highlighting its generality and effectiveness in mitigating multi-image hallucinations.}

\label{main_exper}
\begin{tabular}{lccccc}
\toprule
\textbf{MODEL} & \textbf{ACCURACY} & \textbf{PRECISION} & \textbf{RECALL} & \textbf{F1 SCORE} & \textbf{YES RATIO} \\
\midrule
\multicolumn{6}{c}{\textbf{Existence}} \\
\midrule
Qwen2.5-VL & 71.59 & 88.69 & 51.67 & 64.70 & 30.08 \\
Qwen2.5-VL + OURS & \textbf{73.25} & \textbf{88.81} & \textbf{55.33} & \textbf{67.66} & \textbf{32.09} \\
\hdashline
Mantis & 63.67 & 89.36 & 31.67 & 46.67 & 18.00 \\
Mantis + OURS & \textbf{64.13} & \textbf{89.47} & \textbf{32.83} & \textbf{47.9} & \textbf{18.71} \\
\hdashline
InternVL2.5 & 74.00 & 88.29 & 57.42 & 69.16 & 33.42 \\
InternVL2.5 + OURS & \textbf{74.92} & 87.79 & \textbf{60.25} & \textbf{71.02} & \textbf{35.34} \\
\hdashline
LLaVA-NeXT-Interleave & 75.75 & 87.97 & 62.92 & 72.68 & 37.17 \\
LLaVA-NeXT-Interleave + OURS & \textbf{76.13} & 87.80 & \textbf{63.92} & \textbf{73.33} & \textbf{37.79} \\
\midrule
\multicolumn{6}{c}{\textbf{Count}} \\
\midrule
Qwen2.5-VL & 57.75 & 86.90 & 18.25 & 30.17 & 10.50 \\
Qwen2.5-VL + OURS & \textbf{58.88} & \textbf{89.88} & \textbf{20.00} & \textbf{32.72} & \textbf{11.12} \\
\hdashline
Mantis & 52.38 & 64.18 & 10.75 & 18.42 & 8.38 \\
Mantis + OURS & \textbf{52.88} & \textbf{64.56} & \textbf{12.75} & \textbf{21.29} & \textbf{9.88} \\
\hdashline
InternVL2.5 & 53.13 & 75.51 & 9.25 & 16.48 & 6.13 \\
InternVL2.5 + OURS & \textbf{53.13} & 72.73 & \textbf{10.00} & \textbf{17.58} & \textbf{6.88} \\
\hdashline
LLaVA-NeXT-Interleave & 55.13 & 73.56 & 16.00 & 26.28 & 10.88 \\
LLaVA-NeXT-Interleave + OURS & \textbf{55.38} & \textbf{74.16} & \textbf{16.50} & \textbf{26.99} & \textbf{11.13} \\
\midrule
\multicolumn{6}{c}{\textbf{Id Consitency}} \\
\midrule
Qwen2.5-VL & 68.75 & 64.88 & 81.75 & 72.35 & 63.00 \\
Qwen2.5-VL + OURS & \textbf{70.75} & \textbf{67.08} & 81.50 & \textbf{73.59} & 60.75 \\
\hdashline
Mantis & 62.63 & 58.32 & 89.50 & 70.31 & 75.88 \\
Mantis + OURS & 62.63 & 58.21 & \textbf{89.50} & \textbf{70.54} & \textbf{76.88} \\
\hdashline
InternVL2.5 & 71.38 & 66.67 & 85.50 & 74.92 & 64.13 \\
InternVL2.5 + OURS & \textbf{73.38} & \textbf{68.51} & \textbf{86.50} & \textbf{76.46} & 63.13 \\
\hdashline
LLaVA-NeXT-Interleave & 51.88 & 50.97 & 99.00 & 67.29 & 97.13 \\
LLaVA-NeXT-Interleave + OURS & \textbf{55.25} & \textbf{51.16} & \textbf{99.00} & \textbf{67.46} & 96.75 \\
\bottomrule
\end{tabular}
\end{table*}

\begin{figure*}[t]
  \centering
  \includegraphics[width=\textwidth]{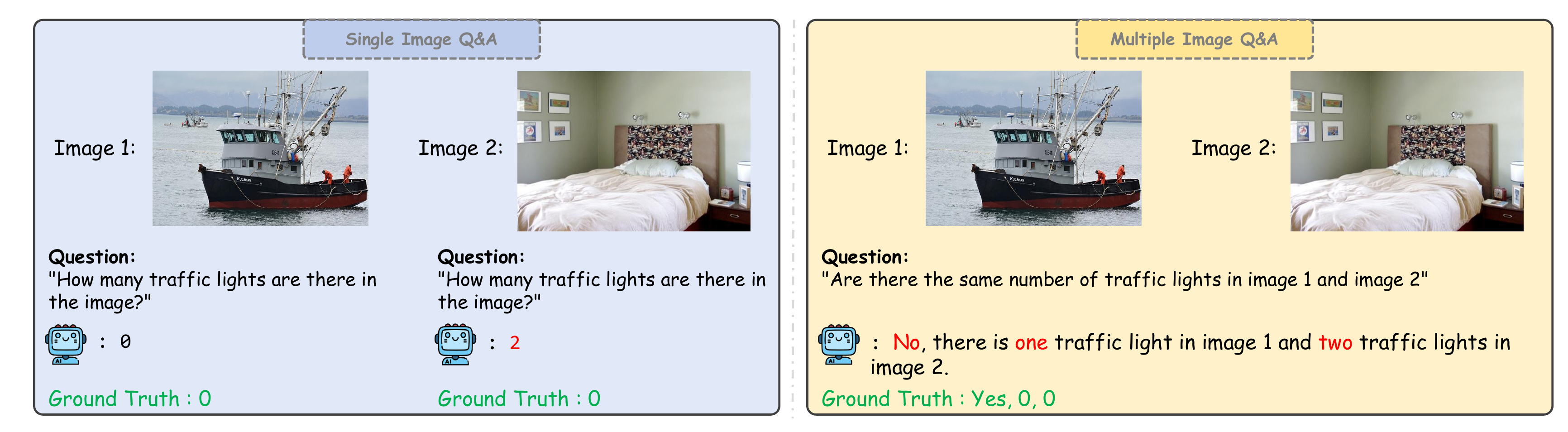}
  \caption{This figure shows how the model makes errors when analyzing images. When asked about each image separately, it correctly sees no traffic lights in Image 1 but wrongly identifies two in Image 2. When comparing both images together, it makes a new mistake by claiming Image 1 has one traffic light while still seeing two in Image 2. This demonstrates that comparing multiple images can introduce new inconsistencies in the model's visual reasoning.}
  \label{single_multi}
\end{figure*}

\section{Experiments}



\subsection{Evaluation Results on MIHBench}

To evaluate the prevalence of multi-image hallucinations and verify the effectiveness of the proposed method, we conduct a comparative analysis across three subtasks: object existence hallucination, object count hallucination, and identity consistency hallucination.

As shown in Table~\ref{main_exper}, object count hallucination proves the most challenging, with consistently low recall and F1 scores, highlighting the complexity of this task and the need for accurate vision-language alignment. Object existence hallucination follows, with moderate model performance under adversarial conditions, while identity consistency hallucination shows the least severe hallucination effects. Among the models, MANTIS demonstrates the weakest performance across all tasks, particularly in accuracy and F1 score, suggesting high vulnerability to hallucinations. In contrast, InternVL2.5 and LLaVA-NeXT-Interleave achieve stronger baseline results. The proposed method consistently improves performance across all models and subtasks, with particularly notable gains in the object count task and further enhancements in identity consistency. These results underscore the method’s effectiveness and generalizability in mitigating multi-image hallucinations. Hallucination examples for each task and the effects of our method will be presented in the supplementary materials.

\subsection{Evaluation of DAB on General Multi-Image Understanding Benchmarks}
To validate the effectiveness of the DAB method, we fix the attention bias coefficient $\alpha$ at 0.5 and evaluate on several non-video multi-image tasks from three general multi-image understanding benchmarks: MMIU~\cite{mmiu}, Muirbench~\cite{muirbench}, and MIRB~\cite{mirb}. The average accuracy across the experiments is summarized in Table \ref{general bench}. As shown in the table, DAB consistently improves performance across all metrics for two distinct multi-image models. The results further demonstrate that DAB consistently improves model performance and its effectiveness and robustness in multi-image understanding.


\begin{figure}[t]
  \centering
  \includegraphics[width=\linewidth]{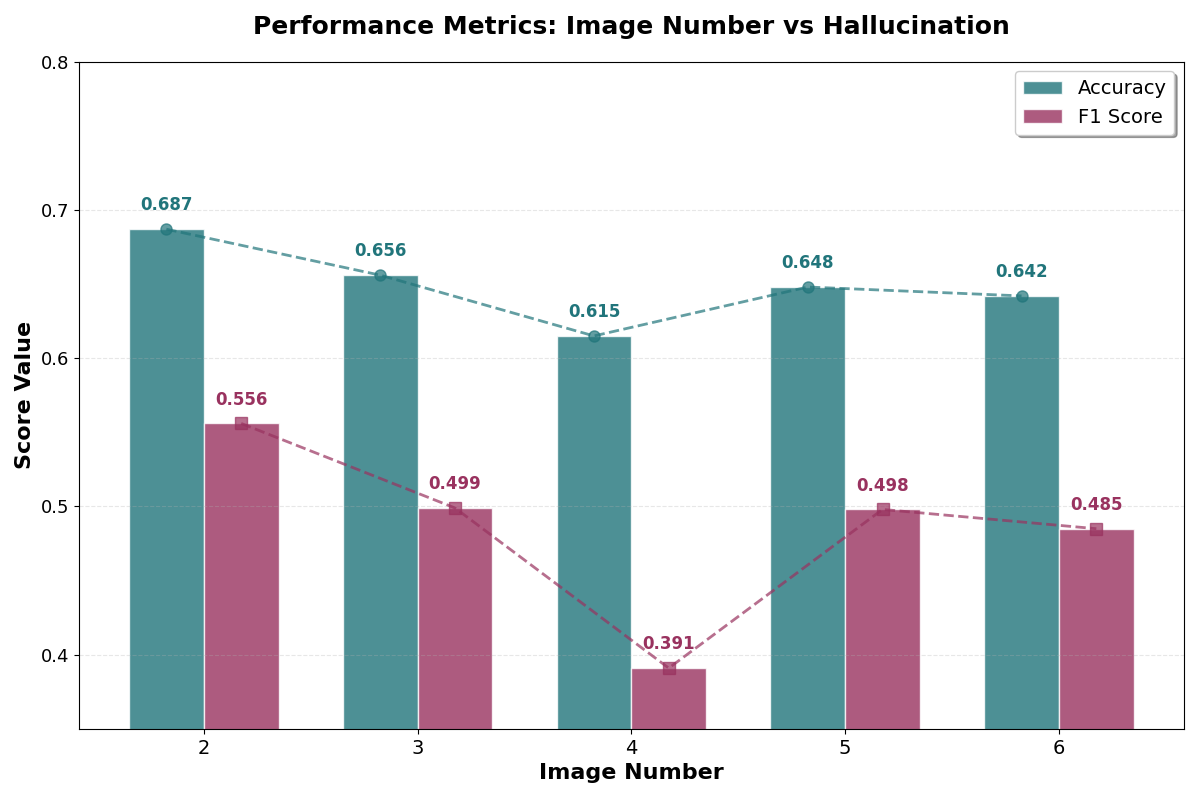}
  \caption{Model performance decreases as the number of input images increases in the task of object existence hallucination. The plot shows accuracy and F1 score declining as image sequence length grows, indicating that longer input image sequences make hallucination.}
  \label{img_lenth}
\end{figure}

\subsection{Causes and Analysis of Hallucination}

Based on observations of the model’s hallucination tendencies in prior single-image tasks, as well as empirical outputs from multi-image scenarios, we observe a notable increase in hallucination frequency when the model processes multiple images. These findings motivate the following hypothesis: the number of image inputs, the presence of single-image hallucinations within the model itself, and the proportion and spatial distribution of negative samples are 

\begin{figure}[!ht]
  \centering
  \includegraphics[width=\linewidth]{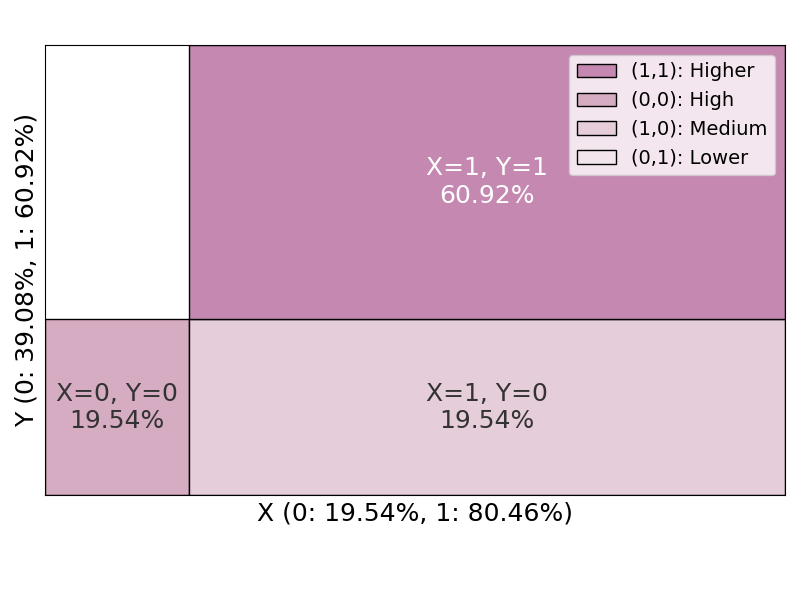}
  \caption{Mosaic plot illustrating how hallucinations occur across single-image and sibling-image sub-questions in multi-image VQA tasks. The variable \(X=1\) indicates hallucinations in any single-image sub-question, while \(Y=1\) indicates hallucinations also appear in sibling images. The dominant \((X=1, Y=1)\) region (60.92\%) suggests that hallucinations often co-occur across related sub-questions. The near absence of \((X=0, Y=1)\) indicates such cases rarely arise without initial hallucinations, reflecting a strong positive correlation between \(X\) and \(Y\).}
  \label{mosaic}
\end{figure}

key factors contributing to the emergence of multi-image hallucinations. In this section, we design targeted experiments across the three tasks of MIHBench to empirically validate this hypothesis. Unless otherwise stated, all analyses are conducted using the Mantis model.

\begin{figure}[t]
  \centering
  \includegraphics[width=\linewidth]{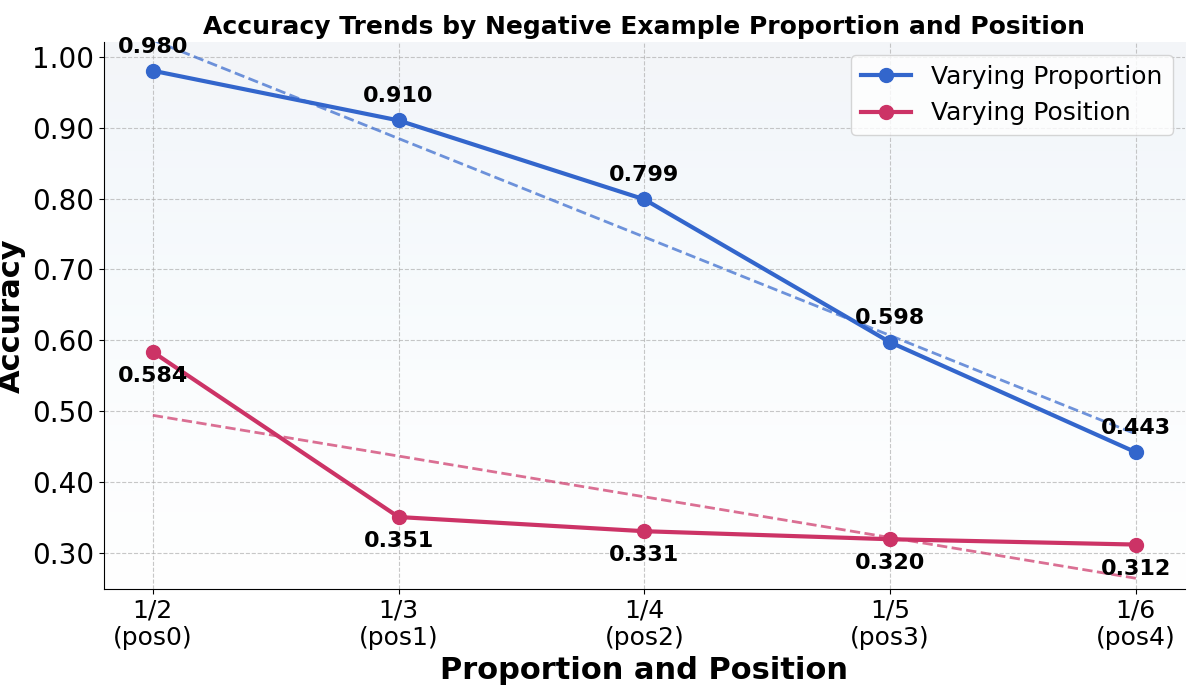}
  \caption{The performance decreases as the proportion of negative examples becomes smaller or as the negative example appears later in the input sequence on the task of identity consistency hallucination. This suggests that when negative evidence is scarce or appears later in the context, the model is more prone to hallucinations.}
  \label{ratio_pos}
\end{figure}

\noindent
\textbf{Number of Images} To investigate the impact of the number of input images on hallucination severity, we extend the original Multi-Image Object Existence Hallucination dataset by constructing subsets with sequence lengths ranging from 2 to 6. For each length, we generate 800 balanced queries, with the negative image consistently placed at the end to mitigate positional bias. As shown in Figure~\ref{img_lenth}, performance generally declines with more input images, \textbf{while the degradation is not strictly monotonic}—indicating that multi-image hallucination severity increases with greater visual input.

\noindent
\textbf{The Correlation Between Hallucinations in Single-Image and Multi-Image Scenarios} We hypothesis that single-image hallucinations may propagate and amplify during multi-image reasoning, so we investigate their potential correlation within the multi-image object count hallucination task. Each original multi-image query is decomposed into two single-image sub-queries, prompting the model to predict the object count for each image separately. These predictions are compared to the model’s original multi-image response. As shown in Figure~\ref{single_multi}, hallucinations are more prevalent in the multi-image setting, suggesting amplification effects during joint reasoning.

To further validate this, we extract the model’s perceived object count per image from the original multi-image responses using Qwen2.5-14B-Instruct~\cite{qwen2.5llm}, and prompt the model separately for each image. We then define two binary random variables, \( X \) and \( Y \), as follows:
\begin{itemize}
    \item \( X \) denotes whether there exists a single-image hallucination in the sub-questions derived from a multi-image problem. If any sub-question contains a hallucination, \( X = 1 \); otherwise, \( X = 0 \). The positions of hallucinated sub-questions are also recorded and stored for subsequent evaluation.
    \item \( Y \) denotes whether, under the condition that \( X = 1 \), the sibling image(s) in the original multi-image response also suffered from hallucinations. \( Y = 1 \) if hallucinations occurred; otherwise, \( Y = 0 \).
\end{itemize}

The empirical joint distribution of \( X \) and \( Y \) as shown in Figure~\ref{mosaic}, cases where both \( X \) and \( Y \) share the same value account for over 80\% of the data, and the related Pearson correlation coefficient is 0.6153, reinforcing the conclusion that single-image hallucinations are strongly associated with multi-image hallucinations.

\noindent
\textbf{Proportion of Negative Image} Fixing negative examples at the sequence start while increasing positive images per query systematically reduces the negative ratio and re-evaluates hallucination behavior. As shown in Figure~\ref{ratio_pos}, a higher proportion of positive images biases the model toward assuming object identity consistency across inputs, hindering negative case detection and increasing multi-image hallucination risk where dissimilar instances are incorrectly matched.

\noindent
\textbf{Position of Negative Image} We further investigate the impact of the positional placement of negative sample images within the input sequence on the occurrence of hallucinations. In the object identity consistency hallucination task, we systematically fix the position of the negative sample from the first to the last image in the sequence. As illustrated in Figure~\ref{ratio_pos}, our results indicate that hallucinations are more likely to occur when the negative image appears later in the sequence. In such cases, the model tends to overlook semantic inconsistencies, leading to incorrect consistency judgments and triggering multi-image hallucinations.

\begin{table}[!tbp]
    \centering
    \caption{The DAB performance on other general multi-image benchmarks. The results demonstrate that the DAB method enhances model performance on general multi-image tasks.}
    \label{general bench}
    \begin{tabular}{l c c c}
        \toprule
        Models & MMIU & Muirbench & MIRB \\
        \midrule
        Mantis & 0.366 & 0.314 & 0.538 \\
        Mantis+ours & \textbf{0.376} & \textbf{0.327} & \textbf{0.544} \\
        LLaVA-NeXT-Interleave & 0.360 & 0.426 & 0.228 \\
        LLaVA-NeXT-Interleave+ours & \textbf{0.374} & \textbf{0.453} & \textbf{0.232} \\
        \bottomrule
    \end{tabular}
\end{table}

\section{Limitation}
Although MIHBench and the DAB mechanism advance multi-image hallucination research, several limitations remain. First, the benchmark—built on datasets like MSCOCO 2014 and CO3D—may not fully reflect real-world complexity, limiting external validity. Second, MIHBench primarily evaluates object existence, count, and identity consistency hallucinations, but does not address finer-grained types such as relational or attribute-level inconsistencies. Finally, DAB relies on an empirically set balancing coefficient ($\alpha = 0.5$), and its robustness across architectures and data distributions requires further study.

\section{Conclusion}
We present \textbf{MIHBench}, the first benchmark tailored to evaluating multi-image hallucinations in MLLMs, covering three core tasks: object existence, count, and identity consistency. Our analysis reveals that hallucination severity increases with the number of input images, often propagates from single-image errors, and is influenced by the proportion and position of negative samples. To mitigate this, we propose \textbf{Dynamic Attention Balancing (DAB)}, a training-free mechanism that adaptively equalizes attention across images during decoding. DAB significantly reduces hallucinations and improves performance across all MIHBench tasks and models, demonstrating effectiveness and generalizability. This work provides new insights and tools for understanding and mitigating multi-image hallucinations in MLLMs.

\begin{acks}
This work was supported by National Key R\&D Program of China (No.2023YFB4502804), the National Science Fund for Distinguished Young Scholars (No.62025603), the National Natural Science Foundation of China (No. U22B2051, No. U21B2037 , No. 62302411), China Postdoctoral Science Foundation (No. 2023M732948), and the Zhongguancun Academy, Beijing, China (No. 20240103).
\end{acks}



\bibliographystyle{ACM-Reference-Format}

\end{document}